\documentclass[letterpaper, 10 pt, conference]{IEEEtran}

\usepackage{subcaption}

\DeclareCaptionLabelSeparator{periodspace}{.\quad}
\usepackage[dvipsnames]{xcolor}
\usepackage[normalem]{ulem}

\usepackage[ruled,noend,linesnumbered]{algorithm2e}
\usepackage{float}
\usepackage[pagebackref=true,breaklinks=true,colorlinks,bookmarks=false]{hyperref}

\usepackage{xr-hyper}

\usepackage{tikz}
\usepackage{textcomp}
\usepackage{lipsum}
\usepackage{footnote}  
\usepackage{booktabs}

\usepackage{array}
\usepackage{makecell, caption}

\usepackage{bbm}
\usepackage{amssymb}
\usepackage{amsmath}
\usepackage{multirow}
\newcommand{\ignore}[1]{}

\newcommand{\argmax}[1]{\underset{#1}{\operatorname{arg}\,\operatorname{max}}\;}

\IEEEoverridecommandlockouts                              % This command is only needed if 
\title{\LARGE \bf
AGPNet - Autonomous Grading Policy Network
}

\author{Chana Ross$^{1}$\textsuperscript{*}, Yakov Miron$^{1}$\textsuperscript{*}, Yuval Goldfracht$^{1}$\textsuperscript{\textdagger}$^{2}$, Dotan Di Castro$^1$\\ 
\thanks{$^1$Bosch Center for Artificial Intelligence, Haifa, Israel} \thanks{$^2$Charney School of Marine Sciences, University of Haifa, Israel} 
\thanks{\textsuperscript{*}Equal Contribution}
\thanks{\textsuperscript{\textdagger}This work was done during an internship at BCAI}

{\tt\small {\{Chana.Ross, Yakov.Miron, Dotan.DiCastro\}}@bosch.com}, 
{\tt\small Yuval.Goldfracht@gmail.com}

}

\renewcommand\footnotemark{}

\begin{document}
\makeatletter
\setcounter{figure}{-2}
\g@addto@macro\@maketitle{
  \begin{figure}[H]
  \setlength{\linewidth}{\textwidth}
  \setlength{\hsize}{\textwidth}
    \centering

  \includegraphics[width=0.774\linewidth, height=8cm]{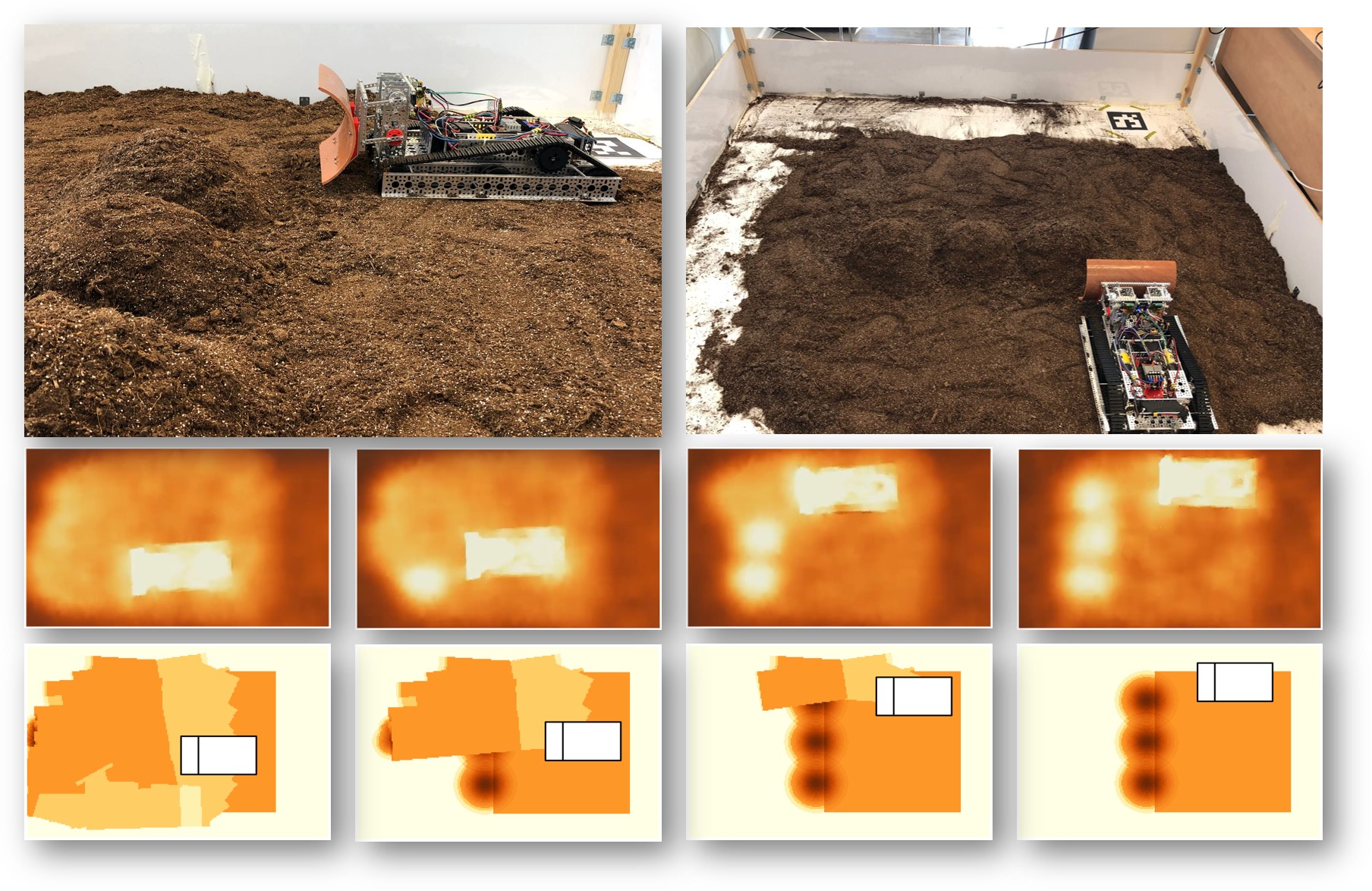}
  
  \caption{Comparison of the ability of a \textbf{simulated environment} and a \textbf{real dozer} to grade an area studded with piles. \textbf{Top row:} RGB images of our experimental setup  (see Section \ref{Real_world_comparison}) showing the scaled dozer facing the sand piles. \textbf{Middle row:} Representative height-maps of states in the real environment. Depth observations were projected onto world coordinates using orthographic projections \cite{MVG}. \textbf{Bottom row:} Similar states observed in the simulated environment. In the middle and bottom rows, the right column is the initial state space, and all the others, states after actions were taken in the simulated and real environments. This figure clearly depicts the resemblance between the simulated and real height-maps in the grading task.}
    \label{fig:sim2real_all}
  \end{figure}
}

\maketitle
\thispagestyle{plain}
\pagestyle{plain}

\begin{abstract}
% In this work, we establish heuristics and learning strategies for the autonomous control of a dozer grading an uneven area studded with sand piles. Our approach draws from reinforcement learning, behavior cloning and contrastive learning methods. We formalize the problem as a reinforcement learning task, design and demonstrate agent-environment interactions in a simulation and finally show our results on a prototype. After training an autonomous agent, we conduct extensive experiments in an approximated simulation and on a real dozer. Our model combines several methods, reaches human-level performance and outperforms current state-of-the-art machine learning methods for the autonomous grading task in terms of total reward.

In this work, we establish heuristics and learning strategies for the autonomous control of a dozer grading an uneven area studded with sand piles. We formalize the problem as a Markov Decision Process, design a simulation which demonstrates agent-environment interactions and finally compare our simulator to a real dozer prototype. We use methods from reinforcement learning, behavior cloning and contrastive learning to train a hybrid policy. Our trained agent, AGPNet, reaches human-level performance and outperforms current state-of-the-art machine learning methods for the autonomous grading task. In addition, our agent is capable of generalizing from random scenarios to unseen real world problems.
\end{abstract}

\section{Introduction} \label{Introduction}

% \externaldocument[V1-]{appendix} 

In this work, we focus on autonomous path planning for construction site vehicles. Specifically, we discuss the task of grading a given area with a number of sand piles. In this task, the dozer is confronted with an uneven terrain (Fig. \ref{fig:sim2real_all}) and is required to level the ground, in a minimal amount of time, to a predefined target height. Problems of this sort in the off-road autonomous driving industry have attracted increasing interest in the past two decades due to shortage in experienced drivers as well as risings demands in the construction industry. Previous work has mainly focused on obstacle avoidance \cite{Kelly_Alonzo, xiang2021extension}, optimal trajectory planning \cite{zhang2018integrating, Swift2019PathPF, xiang2021extension} and traversability \cite{Zhu_2020}. 

The autonomous grading task was first tackled from a path-planning perspective by \cite{hirayama2019path}. This pioneering study was the first to directly address the autonomous grading problem. In their work, \cite{hirayama2019path} implemented a rule-based approach, whereby, given a large sand pile, the system selects the goal points the agent needs to reach and the grading leg it needs to perform. After the agent aggregates several of these legs, the pile is graded and the task is considered done. 
% They also compared several rule-based approaches for these legs, e.g., full blade capacity, half capacity, etc. (the volume of the sand on the blade directly effects the dozer's velocity and, therefore, the duration of the assignment).
While this approach relies on rule-based heuristics, recent successes with machine learning methods have demonstrated the possibility of automating and optimizing such complex problems.

In this work, we have chosen to tackle the problem of trajectory planning using reinforcement learning (RL), a sub-field of machine learning. We combined and compared various techniques, such as behavior cloning and contrastive learning. We also utilized domain knowledge in order to design an appropriate action space and add a prior to the initial action distribution so that it is easier for the agent to learn (see \ref{Gaussian_masking} for an example of such a prior). 

As we demonstrate herein, our model  is capable of training on a random scene, can generalize for a more complex problem and preforms well on all the realistic scenes tested. To validate our method, we created a simulation for training and evaluating our models, which 
includes all the important interactions between the dozer and the soil. 
% encode the important features when an agent interacts with soil.
In addition, we built a lab setup for small-scale algorithm testing (see  \ref{Real_world_comparison}).

Our main contributions are: 
\textbf(1) We provide an end-to-end pipeline for training autonomous dozers that combines supervised learning and RL for improved robustness, enhanced performance and reduced sample complexity. Here, we implement a hierarchical architecture in which high-level trajectory planning is  learnt and low-level action control is performed.
\textbf(2) We establish an RL environment simulator for the earth-moving dynamics and the interaction between the dozer and the soil. Using this simulator, we train an RL agent for the autonomous-grading task.
(3) We validate the simulator using a real prototype dozer and compare height maps generated by our simulator to those taken by a real depth camera.

%\cite{https://doi.org/10.1002/rob.20147}

\section{Related Work} \label{Related Work}
Algorithms that enable real-world navigation and trajectory planning for autonomous vehicles have been extensively studied in many real-world domains, such as outdoor driving \cite{doi:10.1177/0278364908090949}, areal drones \cite{4472379}, \cite{5980357} and indoor navigation \cite{rosen1967application}, \cite{Wallace1985FirstRI}, \cite{Furgale2010VisualTA}. 
The  predominant  approach  for  autonomous vehicles includes building a map, localizing the agent in the map and using the map for planning \cite{DBLP:journals/corr/abs-1909-05214}.
While these approaches have enabled recent state-of-the-art results \cite{DBLP:journals/corr/abs-1912-04838}, they still face a number of open challenges in other domains, such as construction vehicles and off-road driving.

Unlike on-road driving, early works in the off-road domain show limited performance \cite{Kelly_Alonzo}. They typically offer a system architecture that includes perception, mapping, RT motion planning and some software implementation guidelines. Their sensors include cameras, Lidars, GNSS systems and fusion algorithms as inputs for the motion planner. Their motion planner relies on a hierarchical approach that utilizes basic rule-based search algorithms (A* for example). \cite{Zhu_2020} suggest learning optimal trajectories from the demonstration (LFD) of an experienced driver using Deep Maximum Entropy Inverse  RL. They tested their approach on a jeep vehicle in the context of traversability analysis. Moreover, \cite{zhang2018integrating} integrated the kinematics of the vehicle for enhanced robustness. All the above approaches mitigate the problem of trajectory planing in the context of obstacle avoidance while including the most minimal interaction with the environment possible.

We focus our research on the path-planning algorithm for autonomous earth-moving vehicles. The most successful algorithms for these problems use classic solutions similar to \cite{hirayama2019path}, \cite{Swift2019PathPF}, \cite{Kim2019Developing}, \cite{xiang2021extension}. 
In \cite{hirayama2019path}, the authors implemented a heuristic approach to reach the piles and examined the trade-off between increasing the blade's grading of the pile to full capacity and pushing less sand to reduce the elapsed time and sand spillage. \cite{Swift2019PathPF} used graphs to describe the locations of sand piles and locations not feasible for sand. By combining these graphs and using basic rule-based logic, they found the optimal path for moving multiple objects to multiple goals. 
Finally, \cite{xiang2021extension} attempted to solve the multi-agent problem by using graphs to describe the environment and agents in the scene and simple optimizations to determine the location of each agent.

Recent work using learning techniques such as RL \cite{IEEE_excavation}, \cite{9004935}, behaviour cloning (BC) \cite{DBLP:journals/corr/abs-2010-04767} and imitation learning (IL) \cite{Son2020ExpertEmulatingET} attempt to solve path-planning autonomous driving by focusing mainly on low-level control of excavators. All learning algorithms use trial-and-error or supervised-learning mechanisms to learn the interaction between the environment and the vehicles and to optimize the policy. RL assumes a simulation of the interaction or of the real-world vehicle is available for training the agent, whereas IL and BC assume an experienced driver data set exists.

To the best of our knowledge, very little research has been done on bulldozers, specifically regarding path-planning optimization using learning methods. This can be attributed to the presumed simplicity of the problem. However, the actual operation of bulldozers requires skillful techniques because the behavior of the soil is irreversible \cite{6030650}. \cite{9385686}, \cite{Nakatani2018AutonomousGW} and \cite{8591189} solved a simplified environment with one sand pile using DQN and PPO. These approaches yielded optimal results for the simple problems they solved, but it would be hard to expand them to more complex problems.   

Our research attempts to solve more complex scenes and problems by combining a continuous mathematics-based simulation for training RL algorithms with BC and other learning techniques to improve the models' ability to detect important information regarding the state.  

We include in our solution methods that have recently shown great success in other domains, such as contrastive learning \cite{srinivas2020curl}, hierarchical action spaces \cite{DBLP:journals/corr/abs-2005-03863}, hybrid RL and BC algorithms \cite{Booher2019BCR} and action space down-sampling \cite{DBLP:journals/corr/abs-2105-06411}. We show how combining these techniques creates an overall robust solution with optimal results on different real-world scenarios with reduced sample complexity. Finally, to validate our simulation, we built a prototype dozer that captures the important elements needed for training algorithms.

\section{Our Approach: AGPNet} \label{Our Approach}
Our goal is to create an optimal policy for a bulldozer preforming a grading task, where given a target area and number of sand piles, the dozer must find an optimal trajectory to flatten the sand piles to a predetermined target height. 
We solved the problem using a combination of behavioral cloning, contrastive learning and model-free reinforcement learning techniques. In the following section, we describe the problem formulation, learning methodology and full algorithms we used to solve the problem.
%%%%%%%%%%%%%%%%%%%%%%%%%%%%%%%%%
%%%%%%%%%%%%%%%%%%%%%%%%%%%%%%%%%
\subsection{Autonomous Grading: Problem Formulation} \label{Problem Formulation}
% \chen{Problem formulation mixes into it the RL preliminaries and MDP explanation. Why not (1) explain MDPs and RL in the general sense. (2) given the RL formulation, explain how you define your task as an MDP (states, actions, rewards)? The task itself is already defined in the intro, so now you just need to formalize it.}
Given an initial height-map, $H_{init}$, and a desired height-map, $H_{des}$, the agent's goal is to reach $H_{des}$ in minimal time. To simplify the problem, we define at every time point, $t$, the difference between the current height-map, $H_t$, and the desired one as: $\delta_{H_t}= H_{t} - H_{des}$. The sequential reward for each step in the episode is: $r_t = \delta_{H_{t-1}} - \delta_{H_t}$, i.e., the dozer receives a positive reward if $H_t$ is closer to $H_{des}$.

The goal is to maximize the overall reward and minimize the difference between the final height-map and the target:
\newline $\pi^* = \argmax{\pi \in \Pi} [J(\pi)]$,
where $J(\pi) = \mathbb{E}[\sum^T_{t=0}(\gamma^t  r_t)]$ (the ideal policy manages to grade all the sand above the wanted height in each pixel). 
\begin{figure}
 \centering
 \begin{subfigure}[b]{0.15\textwidth}
     \centering
     \includegraphics[width=\textwidth]{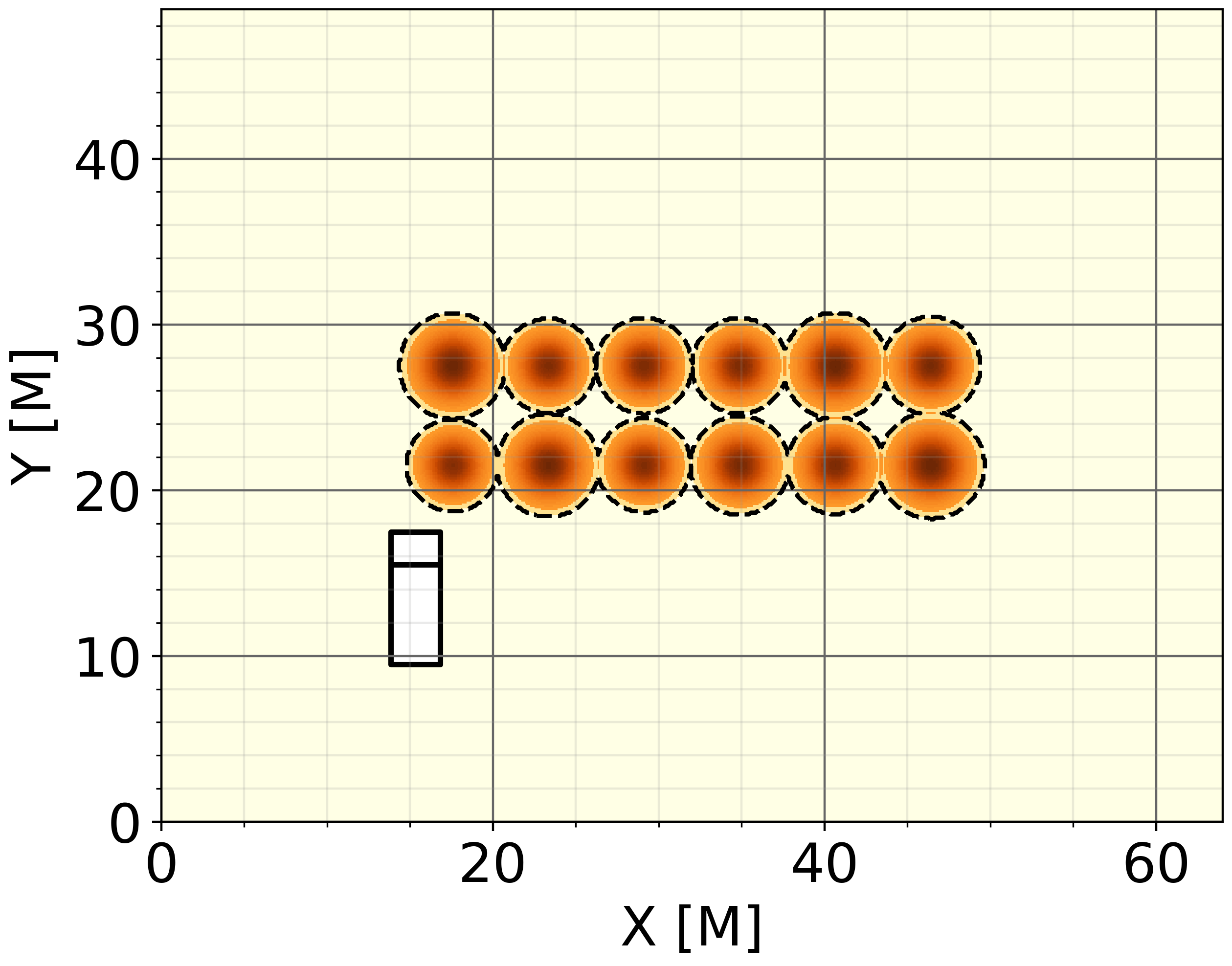}
     \caption{Initialization}
     \label{fig:scenario_init}
 \end{subfigure}
 \hfill
 \begin{subfigure}[b]{0.15\textwidth}
     \centering
     \includegraphics[width=\textwidth]{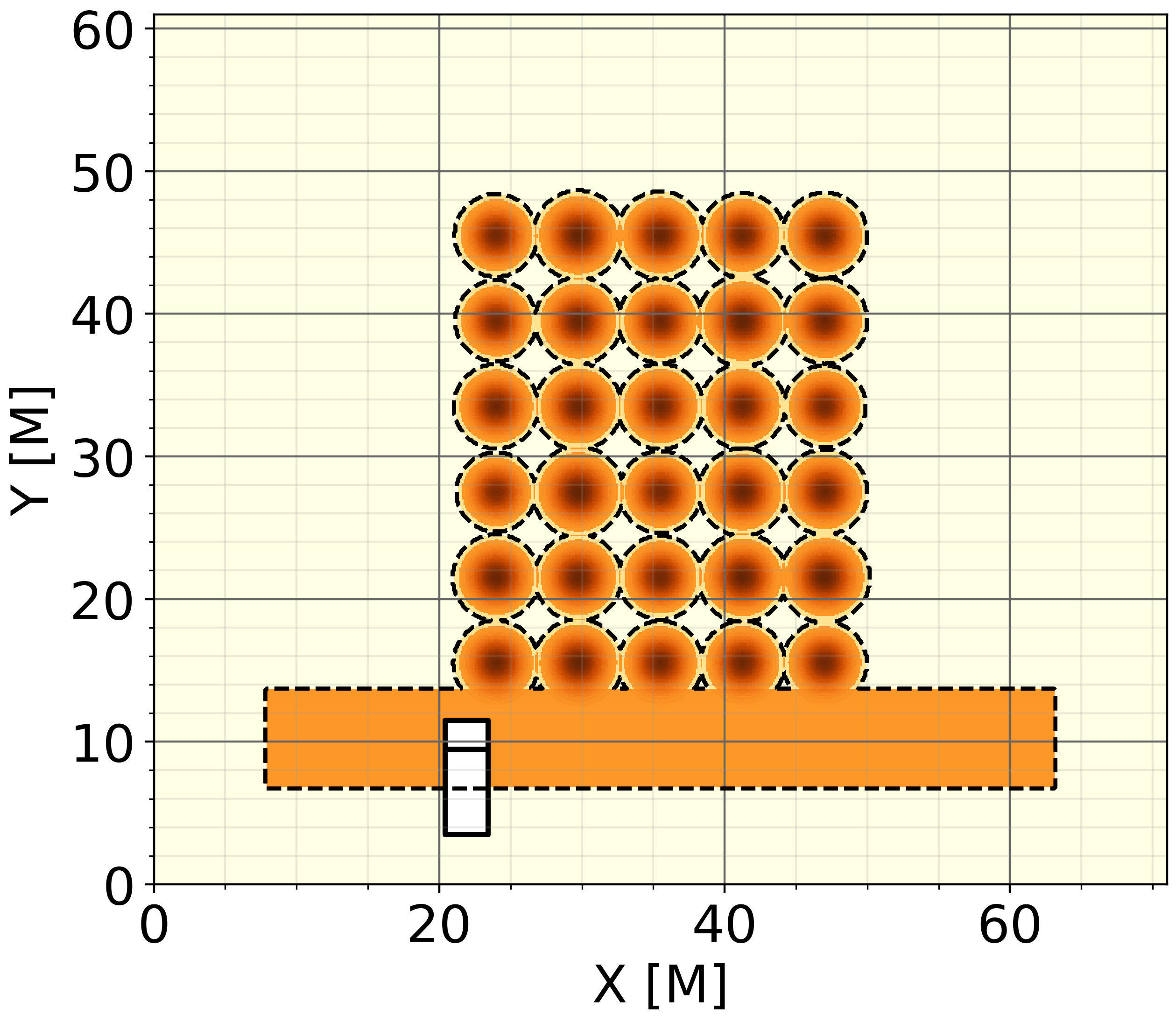}
     \caption{Continuous}
     \label{fig:scenario_cont}
 \end{subfigure}
 \hfill
 \begin{subfigure}[b]{0.15\textwidth}
     \centering
     \includegraphics[width=\textwidth]{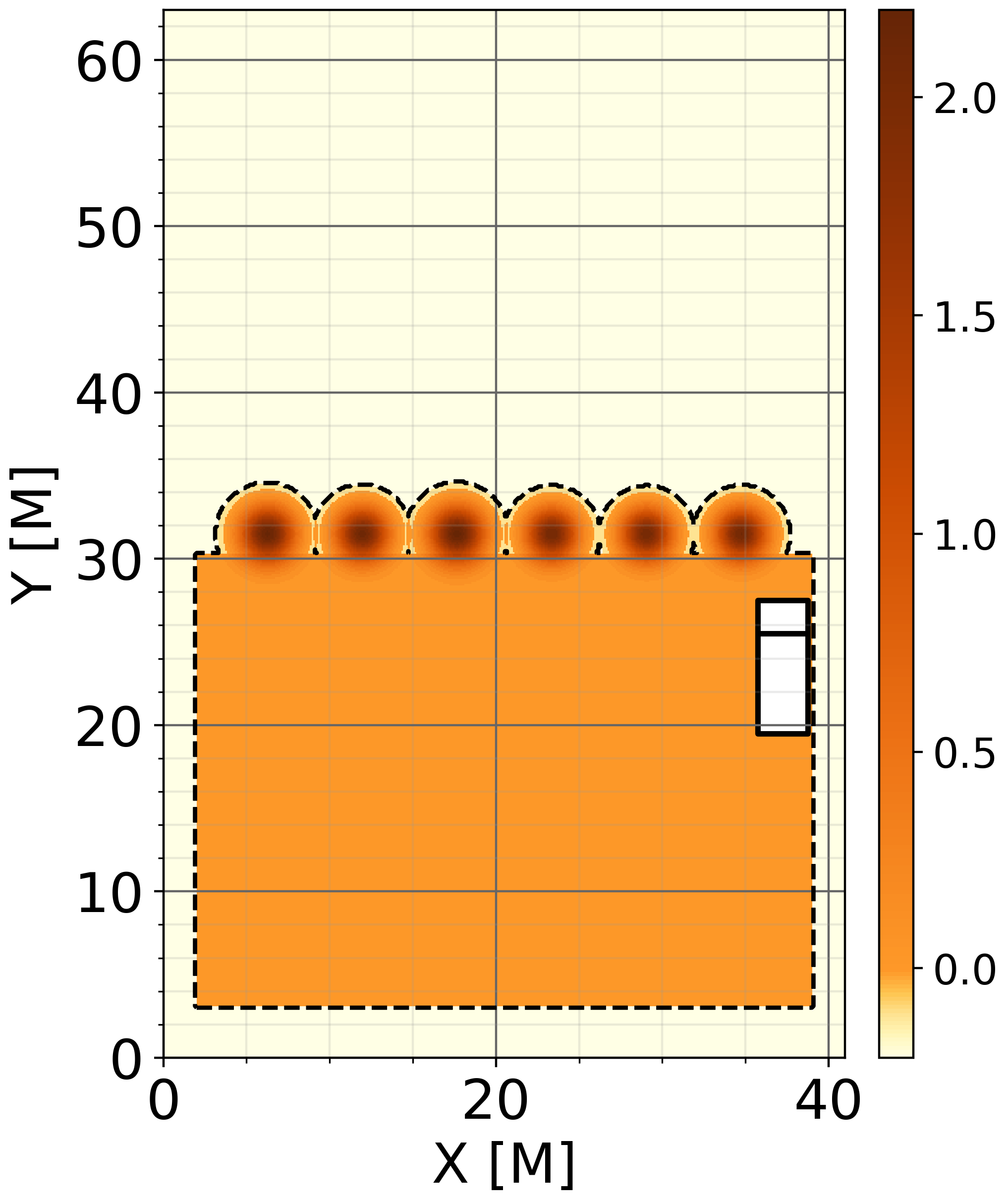}
     \caption{Edge}
     \label{fig:scenario_edge}
 \end{subfigure}
    \caption{The first state of the area for each sub problem. In the \textbf{initial} scenarios, there is no graded area and a few rows of sand are dumped. The dozer needs to create an incline and reach the target height, $H_{des}$, if possible. Meanwhile, the dumper will add more sand piles for grading in front of the initial ones, creating more rows of sand. In the \textbf{continuous} scenarios, the agent is located at $H_{des}$, i.e., on top of the previously graded area, and sand piles are continuously being added to the vicinity of the graded area. The task is to constantly grade them to $H_{des}$. The main difference from the previous problem is that the piles are dumped on top of $H_{des}$ and the dozer needs to push it forward, thus enlarging the area in which $H = H_{des}$. Finally, in the \textbf{edge} scenarios, the final row of sand piles is dumped, most of the area is already graded, and the sand leftovers need to be cleared. The dozer must create a decline to flatten the sand and then smooth the graded area.}
    \label{fig:scenario_examples}
\end{figure}
Given a number of sand piles in a defined area, we can divide the problem into three stages: Initial, Continuous and Edge (see Figure \ref{fig:scenario_examples}).    
We used these sub-problems during inference for our algorithms to test generalization. Of note, despite learning on random scenarios, our agent succeeded in these real-world problems.
%%%%%%%%%%%%%%%%%%%%%%%%%%%%%%%%%
%%%%%%%%%%%%%%%%%%%%%%%%%%%%%%%%%
\subsection{Our problem formulated as an MDP}
We viewed the task as a Markov Decision Process (MDP), which can be defined as a tuple $(S,A,P,\gamma,R)$ where $S$ denotes the state space of the scene, $A$ the action set of the autonomous vehicle, $P$ the state transition probabilities, $\gamma \in [0,1)$ the discount factor, and $R$ the reward for a state.

\begin{figure}
 \centering
 \begin{subfigure}[b]{0.2\textwidth}
     \includegraphics[width=\textwidth]{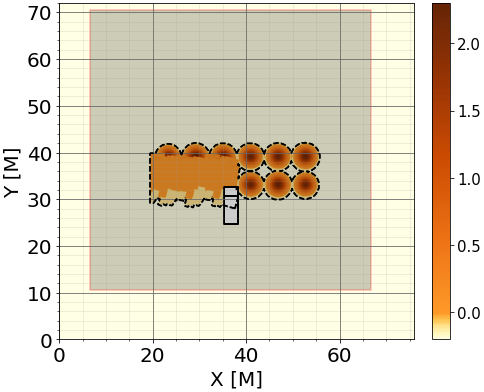}
     \caption{State}
     \label{fig:state_example}
 \end{subfigure}
%  \hfill
 \begin{subfigure}[b]{0.2\textwidth}
     \includegraphics[width=\textwidth]{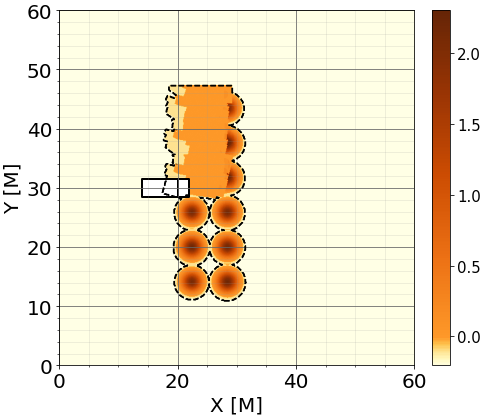}
     \caption{Observation}
     \label{fig:obs_example}
 \end{subfigure}
    \caption{An example of a State (a) and Observation (b) used in the simulation. The shaded area in the state indicates the FOV of the dozer. While the dozer moves in the state, it is stationary in the observation.}
    \label{fig:state_obs}
\end{figure}
\textbf{State}:
The State \textbf{$s_t \in R^{m \times n}$} is the full area's difference height-map, $\delta_{H_t}$ (Figure \ref{fig:state_example}). As the states dimensions may vary and the agent needs to generalize to all domain sizes, the observation size is fixed to be the EGO Field Of View (FOV). This may result in partial observability, Yet, empirically, this was not found to be an issue when the observation was large enough. In addition, we down-sampled (see \ref{down-sampling}) the observation to maintain a small action space. This meant projecting points farther away from the agent in the point cloud to the agents' coordinates, as shown in Figure \ref{fig:obs_example}, where $m\times n$ is the size of the FOV. Note that, for simplicity, hereunder the term \textit{state} refers to the observation. In addition, the actions are defined in the agent's coordinates; therefore, having a state that reflects the action space reduces the complexity during training, as the state space is scale, rotation and translation invariant.

\begin{figure}
 \centering
 \begin{subfigure}[b]{0.15\textwidth}
     \includegraphics[width=\textwidth]{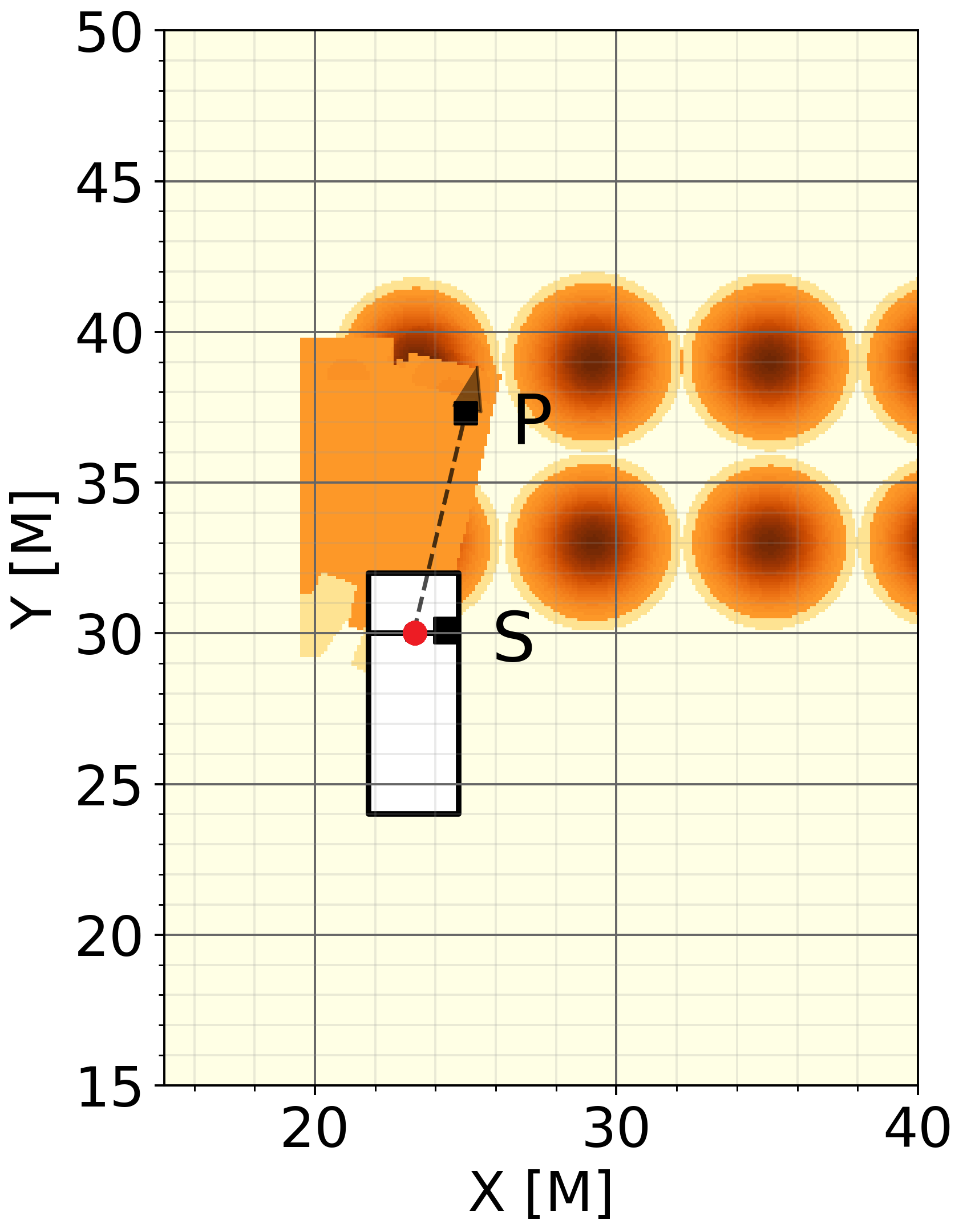}
     \caption{original state}
     \label{fig:snp_exam1}
 \end{subfigure}
%  \hfill
 \begin{subfigure}[b]{0.15\textwidth}
     \includegraphics[width=\textwidth]{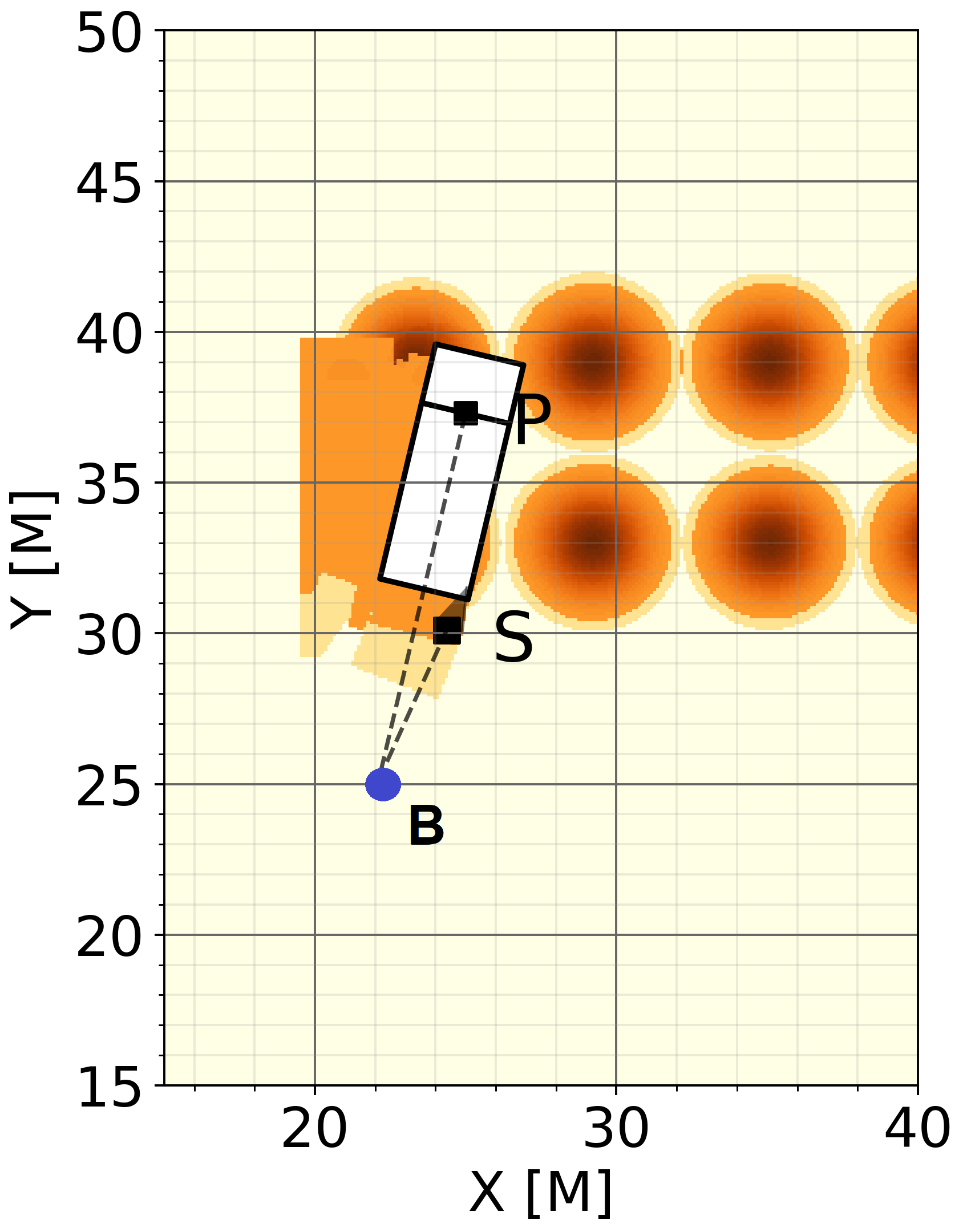}
     \caption{state after $p$}
     \label{fig:snp_exam2}
 \end{subfigure}
 \begin{subfigure}[b]{0.15\textwidth}
     \includegraphics[width=\textwidth]{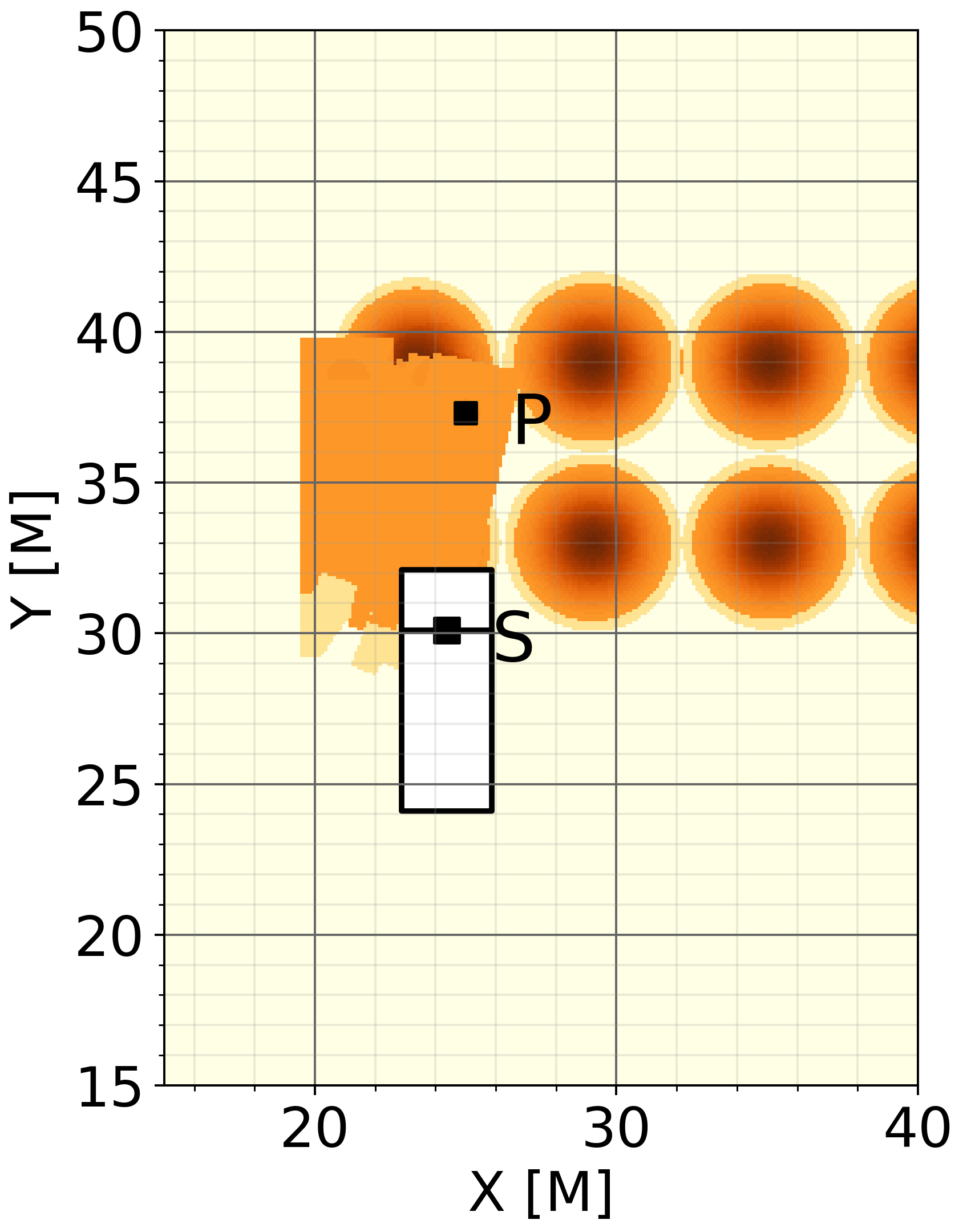}
     \caption{state after $s$}
     \label{fig:snp_exam3}
 \end{subfigure}
    \caption{Example of MDP actions ($p$, $s$) and the trajectory between these actions.
    In \ref{fig:snp_exam1}, the initial position is the red dot. The action's order is as follows:
    (i) From origin rotate to face $p$.
    (ii) Drive forward to $p$. This action has the greatest value w.r.t. the received reward, as it is the only one that grades sand.
    (iii) Reverse back to B (blue dot in \ref{fig:snp_exam2}).
    (iv) Rotate to face next $s$.
    (v) Drive forward to next $s$.
    (vi) Rotate to face the sand piles (\ref{fig:snp_exam3}).}
\label{fig:snp_example}

\end{figure}

\textbf{Action}: To simplify the action space and reduce complexity, we approached the problem in a hierarchical manner by implementing two levels of action: 
    \textit{(i)} \textbf{Way-point planning agent} (high level): pixel level actions. In this case, the action is paramaterized as a tuple $(p_i, s_i)$ and the agent chooses two pixels, where $p_i$ is the first way-point reached by a forward movement and $s_i$ is the starting point of the next action. Once these points are selected, another agent chooses the optimal path and generates the "low-level" actions for the vehicle (see Figure \ref{fig:snp_example}).
    \textit{(ii)} \textbf{Trajectory generation agent} (low level): Here, a trajectory is created  between the way-points chosen by the way-point planning agent. Since a dozer typically drives in straight lines to avoid slippage, we chose to define these actions as a continuous movement in the dozer's body axis: $\delta_x$ for translation and $\delta_\psi$ for rotation.

We focused the RL agent on the high-level planning (way-point) and used the heuristic approach for the low-level actions (similar to \cite{hirayama2019path}, where, given the state and way-points, a path is automatically planned). 

We define $a_t = (p_t, s_t)$ as a two dimensional action containing the destination, $p$, and the next starting point, $s$ (see the way-point planning explanation above). This ensures that prior knowledge from the previous action, $p_t$, can be encoded and embedded in the next action, $s_t$. Combining these actions ($a_t = (p_t, s_t)$) to form one RL action allows the policy to learn the correlation between the sub-actions and choose the correct sequence of trajectories. 
Action \textbf{$a_t$} will be sampled from the two policy distributions over the FOV pixel map of the agent $(\pi \in R^{2 \times \hat{m} \times \hat{n}}, \hat{m} = \frac{m}{2^{N}}, \hat{n} = \frac{n}{2^{N}})$, where $N$ is the down-sampling factor, as depicted in \ref{down-sampling}. 

\textbf{Reward}: The grading task involves many objectives, a feature that needs to be reflected in the reward function. In our environment, the reward is multi-modal. One mode takes time as an objective, so only discounting over the horizon might collapse to the trivial solution (minimal time while not completing the task or not touching the sand at all in the fatal case). In the general case, the optimal agent will complete the task, i.e., reach the target surface, $H_{des}$, not leave sand piles/bumps in the area i.e., will remove the maximum volume, and grade sand in every leg, i.e., will minimize the legs in which reverse/rotation actions are selected. Moreover, upon task completion, the agent gets a large reward, and if not accomplished, the agent receives a large negative reward. See section \ref{oracle_heuristic}.
(An agent might fail, for example, if it selects an action outside of the possible area). The multi-objective reward function is: $R_t = \lambda_v * f_v - \lambda_t * f_t + \lambda_h * f_h + \lambda_d * \mathbbm{1}_{is_{done}} - \lambda_f * \mathbbm{1}_{is_{failed}}$, where $f_v, f_h$ and $f_t$ are the respective functions of the current volume removed, current height removed and time spent on executing the action. All the $f_i$ functions and  $\lambda_i$ coefficients of the specific rewards were tuned during the hyper-parameter search.

%%%%%%%%%%%%%%%%%%%%%%%%%%%%%%%%%
%%%%%%%%%%%%%%%%%%%%%%%%%%%%%%%%%

%%%%%%%%%%%%%%%%%%%%%%%%%%%%%%%%%
%%%%%%%%%%%%%%%%%%%%%%%%%%%%%%%%%
\subsection{From MDP to simulation and real-world comparison} \label{Real_world_comparison}
Construction site problems are unique in that their complexity lies in the interaction between the vehicle and the soil rather than in the uncertainty of the actions or states. The movement of the soil due to a dozer's action is not trivial and can be \textbf{simulated} using different techniques, with each one capturing a different level of the real interaction \cite{10.1007/11861201_46}, \cite{Sauret2014BulldozingOG}, \cite{Kim2019Developing}. We chose to use Multivariate Gaussian Distributions for the sand piles and implemented the interaction between the soil and the dozer numerically for each action chosen. We decided to implement a simple simulation rather than a more complex one to allow for quick training and an easy-to-configure environment. Despite our simulation being computationally inexpensive, it takes into consideration important dozer behaviour such as velocity change due to torque and sand-dozer interactions, which are needed to estimate an agent's optimal behaviour, e.g., Gaussian distributions changing due to dozer movement and soil being pushed by the blade. 

To validate our simulation, we created a \textbf{dozer prototype} at a $1:9$ scale compared to a real dozer. We built a sand box with an RGBD sensor that provides both height-maps and agent locations within a global coordinate system. The prototype dozer interacts with the sand, and the height-maps can be recorded throughout the episode and be post-processed with the simulation of the same scenario. Figure \ref{fig:sim2real_all} (top right image) shows the lab experimental setup with \textit{(i)} a sand box filled with sand prior to grading, \textit{(ii)} the prototype dozer, and \textit{(iii)} the localization system: Three arUco markers are used to localize the dozer and calibrate the camera w.r.t world coordinates. The dozer is controlled by $3$ motors: two control the tracks and the third controls the blade. The bottom two rows in figure \ref{fig:sim2real_all} show the output of the experimental setup.
% \rc{change to (a),(b),... if you want}

%%%%%%%%%%%%%%%%%%%%%%%%%%%%%%%%%
%%%%%%%%%%%%%%%%%%%%%%%%%%%%%%%%%
% \begin{figure}[!hb]
%     \centering
%     \includegraphics[width=7cm, height=4cm]{images/dozer_1.jpeg}
%     \caption{Lab experimental setup. An example of our initial scenario where the dozer faces the sand piles. Three arUco markers are used to localize and calibrate the camera. The dozer is controlled by $3$ motors: $2$ control the tracks and the third controls the blade.}
%     \label{fig:experamental_setup}
% \end{figure}

%%%%%%%%%%%%%%%%%%%%%%%%%%%%%%%%%
%%%%%%%%%%%%%%%%%%%%%%%%%%%%%%%%%
%%%%%%%%%%%%%%%%%%%%%%%%%%%%%%%%%
%%%%%%%%%%%%%%%%%%%%%%%%%%%%%%%%%
\subsection{Oracle heuristic: SnP} \label{oracle_heuristic}

Throughout our work, we used the rule-based policy inspired by 
\cite{hirayama2019path} as an experienced agent and a good baseline comparison. This policy is based on a human expert and mimics the expert's behaviour. As done in the trained policies, we used two levels of action here: The way-point planner chooses the destination point and then the origin ($p_t, s_t$) and the path planner creates $6$ low-level actions, as shown in Figure \ref{fig:snp_example}.
Unlike our learnt algorithms, the high-level actions are chosen using a combination of classical detection algorithms for sand-pile detection and simple search heuristics.
%%%%%%%%%%%%%%%%%%%%%%%%%%%%%%%%%
%%%%%%%%%%%%%%%%%%%%%%%%%%%%%%%%%
\subsection{Training algorithms: Hybrid approach}

By training multiple agents and comparing their performances, we determined that a hybrid approach that combines multiple methods returns the best overall reward. We denote here the different algorithms used in this paper: 
    %%%%%%%%%%%%%%%%%%%%%%%%%%%%%%%%%

\textbf{Behaviour Cloning} \textbf{(BC)}\label{method:bc}. In BC, the policy is matched to the experienced policy using classic supervised learning algorithms.
In our case, we used the $SnP$ heuristic as our experienced behaviour policy $\pi_B$ and created a data set of episodes, where each example consists of a state-action tuple ($D_B=\{(S_t, a_t)\}$). We trained the policy network $\pi_\theta(a_{t_\theta})$ using a Binary Cross Entropy loss - $L_{BC}(\theta) = \mathbb{L}_{CE}(\pi_\theta(a_{t_\theta}),\pi_B(a_{t_B}))$

    %%%%%%%%%%%%%%%%%%%%%%%%%%%%%%%%%
\textbf{ Online Reinforcement Learning \textbf{(RL)}}\label{method:rl}. 
In online RL, the agent interacts with the environment and finds a policy that maximizes the expected reward for each state. We used policy optimization, where at each iteration, a new set of trajectories is sampled from the simulation using the most updated policy to maximize the policy gradient loss.
Here, in addition to the policy model, we trained the value function $V^\pi_{\theta}(S_t)=\mathbb{E}[J(\pi_{\theta})|S_0=S_t]$ and advantage function $A^{\pi_{\theta}}(s,a)=f(\gamma, R_t)$ (function of $R_t$ at each step). $V^\pi_\theta$ was trained using supervised learning. 
% and $A^{\pi_{\theta_k}}(s,a)$ was calculated after all the trajectories were collected at each learning step.
Specifically, similar to \cite{schulman2017proximal} we trained our networks using the PPO loss. This method has no prior knowledge about the problem, and the policy improves solely based on the simulation reward and is expected to take the longest to train. 
%%%%%%%%%%%%%%%%%%%%%%%%%%%%%%%%%

\textbf{ Contrastive Online RL \textbf{(RL+CL)}}\label{method:cl}. Similar to \cite{srinivas2020curl}, we accelerated the detection capabilities of the RL agent by adding a contrastive unsupervised loss to the training of the embedding layer in the network. As the agent already observes a cropped version of the state (Gaussian masking), we did not utilize cropping for CL. 

    %%%%%%%%%%%%%%%%%%%%%%%%%%%%%%%%%
\textbf{Online RL + residual reward (\textbf{RL+res})}. 
In an attempt to utilize the behaviour policy, on the one hand, and train using on-policy learning and maintain generalization, on the other hand, we used a different RL reward and calculated the residual reward for the current state $S_t$ and action $a_t$ chosen by $\pi_\theta$ (similar to \cite{Johannink_2019}):
$R_t(S_t) = R_{\pi_\theta}(S_t) - R_{\pi_B}(S_t)$, where $R_{\pi_\theta}(S_t)$ is the real simulation reward for the chosen action and $R_{\pi_B}(S_t)$ is the experienced policies' reward in the current state. This method allows on-policy training, on the one hand, and the incorporation of prior knowledge from an experienced driver (oracle policy), on the other hand. Note that this method assumes the experienced policy can be used online during training and that it is not a static data set.

%%%%%%%%%%%%%%%%%%%%%%%%%%%%%%%%%
\textbf{ BC with RL \textbf{(RL+BC)}}.
Unlike \textbf{RL+res}, here, we assume the experienced behaviour policy is not accessible but was collected prior to training or is not easily collected. 
Similar to \cite{Fujimoto2021AMA}, \cite{DBLP:journals/corr/abs-2006-09359} and \cite{Booher2019BCR}, we combined the BC loss with the RL loss:
$L(S, a, \theta_k, \theta)$ = $\lambda_p * L_{PPO} +  \lambda_b * L_{BC}$,
where $\lambda_p$ and $\lambda_b$ are  fixed positive hyper-parameters. For this algorithm, we used episodes from two policies during training: states from the current learnt policy for the RL loss, and states from the experienced policy for the BC loss. This loss is very efficient compared to pure RL, since the agent constantly learns from the behaviour policy in addition to the RL algorithm.

\textbf{RL+BC+CL}. This approach combines RL with BC and CL. Here, we combined the loss or RL and BC, as explained above, and trained the CL loss every $k$ iterations. 

% In Section \ref{Experiments}, we show results for the algorithms listed above and other permutations of these algorithms. 
%%%%%%%%%%%%%%%%%%%%%%%%%%%%%%%%%
%%%%%%%%%%%%%%%%%%%%%%%%%%%%%%%%%

\subsection{Policy Model}
For all the methods, we utilized a deep neural network for estimating the policy for each state. Our models have an embedding layer followed by a number of convolution layers and a fully connected layer at the end. We used Yolo lite \cite{Huang_2018} or Resnet \cite{He_2016} architectures as a baseline and assumed a discrete action space. To ease the learning process, we added a prior to the network such that the agent will initially prefer to move within its near vicinity. This is done by adding a Gaussian layer. Finally, a softmax layer for each sub-action was calculated to produce a distribution over the pixels. An ablation study on the Gaussian distribution parameters and model architecture was done, the results of which are shown in Section \ref{Ablation_study}.

\section{Experiments} \label{Experiments}
To demonstrate our method and the efficacy of our algorithms, we compared them on three types of problems, as outlined in Figure \ref{fig:scenario_examples} and explained in Section \ref{Problem Formulation}. We trained our policy model with $8$ different algorithms and focused our comparison on $4$ metrics. 
The data-set used to train the \textbf{BC} policies included $150$ episodes, each with a range of states drawn from our simulator.
For the purpose of evaluation, we ran $50$ runs for each scenario type generated from the same distribution and compared the mean result for each metric. Each initial state had a different number of sand piles, set up in a lattice format, and the dozer was positioned facing the piles. All the algorithms were calculated on the same scenarios to ensure a fair comparison. Autonomous off-road planning is complex and, specifically, the grading assignment does not have classic solvers for comparison. We, therefore, compared our results to the \textbf{SnP} heuristic that is based on \cite{hirayama2019path}, who used experienced expert drivers and mimicked their behaviour.
\begin{figure}
     \centering
     \begin{subfigure}[b]{0.485\textwidth}
         \centering
         \includegraphics[width=\textwidth]{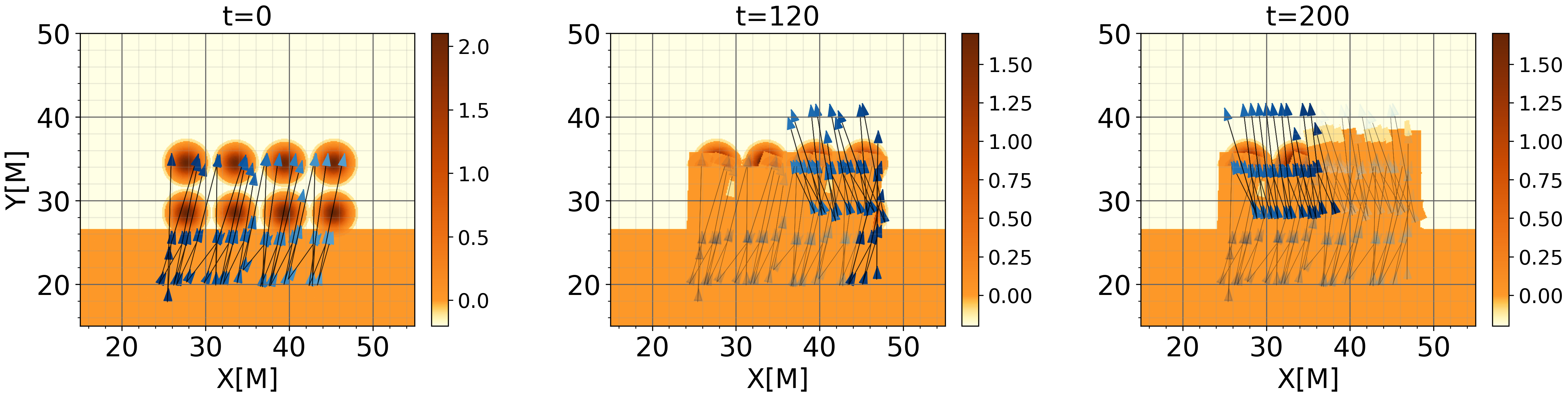}
         \caption{\textbf{SnP, $R_T=1554$}}
         \label{fig:snp_traj}
     \end{subfigure}
     \hfill
     \begin{subfigure}[b]{0.485\textwidth}
         \centering
         \includegraphics[width=\textwidth]{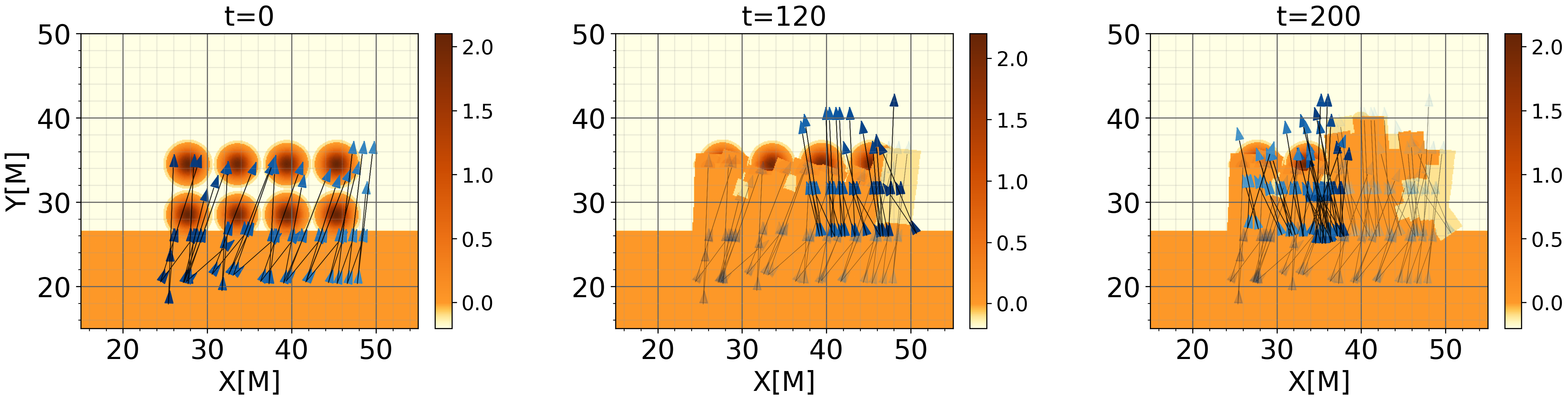}
         \caption{\textbf{BC}, $R_T=1555$}
         \label{fig:bc_traj}
     \end{subfigure}
     \begin{subfigure}[b]{0.485\textwidth}
         \centering
         \includegraphics[width=\textwidth]{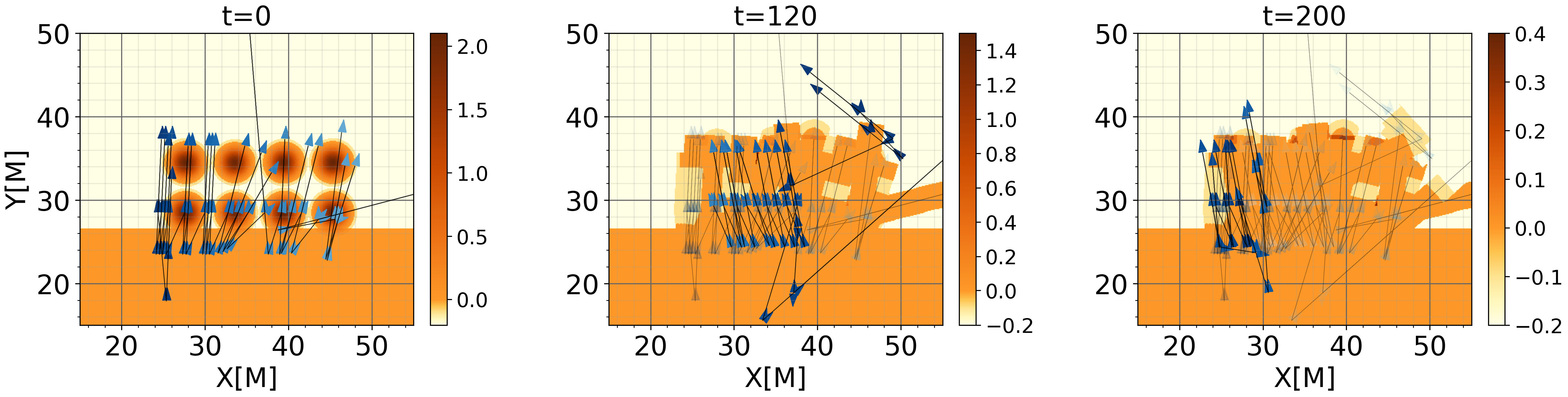}
         \caption{\textbf{RL+BC+CL}, $R_T=1556$}
         \label{fig:rl_traj}
     \end{subfigure}
     \hfill
        \caption{Comparison between trajectories from different policies for the continuous problem. Each triangle is another action and the lines indicate the paths between these actions. Each row shows a different policy (\textbf{SnP}, \textbf{BC}, \textbf{RL+BC+CL}) and each column a different stage of the episode. $R_{T}$ is the total reward.}
        \label{fig:trajectory_example}
\end{figure}
\begin{table*}[ht]
\centering
\resizebox{\textwidth}{!}{%
\begin{tabular}{|c|l|llllllllll|}
\hline
\multicolumn{1}{|l|}{\textbf{scenario type}} &
  \textbf{metric} &
  \textbf{BC} &
  \textbf{BC+CL} &
  \textbf{RL} &
  \textbf{RL+BC} &
  \textbf{RL+BC+CL} &
  \textbf{RL+BC+CL+res} &
  \textbf{RL+BC+res} &
  \textbf{RL+CL} &
  \textbf{RL+res} &
  \textbf{SnP} \\ \hline
\multirow{4}{*}{\textbf{Init}} &
  \textbf{volume $\downarrow$} &
  0.63 &
  21.97 &
  1.03 &
  0.57 &
  0.25 &
  1.49 &
  0.54 &
  \textbf{0.07} &
  37.66 &
  0.36 \\
 &
  \textbf{height $\downarrow$} &
  2.06E-04 &
  7.20E-03 &
  2.71E-04 &
  1.76E-04 &
  \textbf{7.12E-05} &
  4.35E-04 &
  1.52E-04 &
  2.19E-05 &
  9.75E-03 &
  1.15E-04 \\
 &
  \textbf{time $\downarrow$} &
  5024 &
  6566 &
  14464 &
  14326 &
  12975 &
  13778 &
  12263 &
  11140 &
  8166 &
  \textbf{3278} \\
 &
  \textbf{reward $\uparrow$} &
  1478 &
  930 &
  1714 &
  1916 &
  2687 &
  \textbf{3279} &
  2923 &
  3132 &
  741 &
  1934 \\ \hline
\multirow{4}{*}{\textbf{Continuous}} &
  \textbf{volume $\downarrow$} &
  15.04 &
  46.53 &
  0.70 &
  0.36 &
  0.29 &
  0.91 &
  0.74 &
  0.16 &
  23.95 &
  \textbf{0.12} \\
 &
  \textbf{height $\downarrow$} &
  4.10E-03 &
  1.28E-02 &
  1.87E-04 &
  \textbf{9.92E-05} &
  8.63E-05 &
  2.63E-04 &
  2.08E-04 &
  4.98E-05 &
  5.95E-03 &
  3.74E-05 \\
 &
  \textbf{time $\downarrow$} &
  6232 &
  7282 &
  11455 &
  15775 &
  12857 &
  14466 &
  14878 &
  8864 &
  7993 &
  \textbf{3451} \\
 &
  \textbf{reward $\uparrow$} &
  1584 &
  712 &
  1681 &
  2910 &
  2271 &
  \textbf{4101} &
  3365 &
  3123 &
  772 &
  2585 \\ \hline
\multirow{4}{*}{\textbf{Edge}} &
  \textbf{volume $\downarrow$} &
  0.25 &
  4.65 &
  1.22 &
  0.42 &
  0.30 &
  2.16 &
  0.53 &
  0.10 &
  33.37 &
  \textbf{0.05} \\
 &
  \textbf{height $\downarrow$} &
  7.74E-05 &
  1.81E-03 &
  3.49E-04 &
  1.20E-04 &
  \textbf{8.89E-05} &
  6.33E-04 &
  1.53E-04 &
  3.39E-05 &
  8.47E-03 &
  1.42E-05 \\
 &
  \textbf{time $\downarrow$} &
  4657 &
  5670 &
  10981 &
  12065 &
  5993 &
  8333 &
  9802 &
  8468 &
  4633 &
  \textbf{3094} \\
 &
  \textbf{reward $\uparrow$} &
  939 &
  475 &
  1624 &
  2447 &
  \textbf{3690} &
  1901 &
  2528 &
  3512 &
  345 &
  2791 \\ \hline
\end{tabular}%
}
\caption{All the results for our algorithms, including the SnP heuristic, BC, RL and hybrid methods. Our hybrid methods achieve better results on the main metric (height) and the overall reward. Results are mean over 50 i.i.d. runs.}
\label{tab:all_results}
\end{table*}
 
 \begin{figure}[hbt!]
    \centering
    \includegraphics[width=0.72\linewidth]{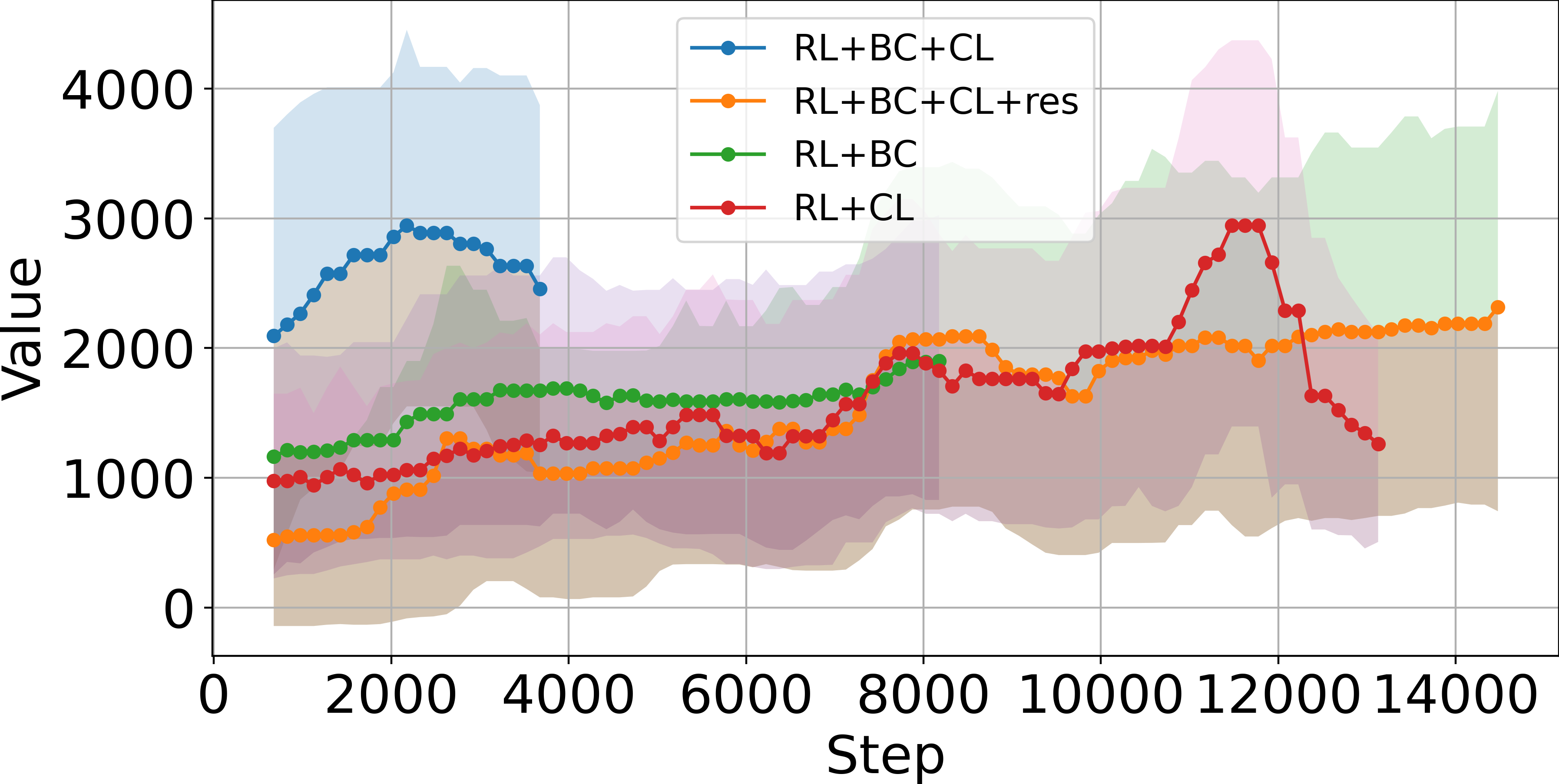}
    \caption{ Example of a RL reward for our top policies during validation on initial scenarios (while training). Here, the mean reward was calculated during training on $50$ different initial episodes. \textbf{RL+BC+CL} reaches the best overall reward \textbf{faster} than other policies. In addition, \textbf{RL+CL} continues to learn throughout all the steps. Note that the std. is large due to a variety of scenarios (each scenario can reach a different reward depending on the initial number of sand piles).}
    \label{fig:rl_loss}
\end{figure}

% \begin{figure}[b]
%     \centering
%     \includegraphics[width=7cm, height=4cm]{images/ablation_reward.png}
%     \caption{total reward graph. in the case $dec_{power} = 3$, the reward is maximal, though the number of parameters is much smaller than $dec_{power} = (1,2)$. }
%     \label{fig:ablation_dec_power}
% \end{figure}

\paragraph{\textbf{AGPNet Results}} 
We show our results on all the metrics in Table \ref{tab:all_results}.
% and specifically focusing on the reward in Figure \ref{fig:algos_results}. 
The results of the majority of our approaches are on par with the baseline heuristic. We found that our combined algorithms approach (RL+BC+CL) outperforms the heuristic in terms of important metrics and overall reward. 
In our approaches, the agent learns from experienced rule-based algorithms similar to other BC models but also trains on online policies allowing for exploration. 
In addition, we enhanced the policies' ability to detect important features in the state space by adding a CL loss (\textbf{CL}), and the final Gaussian masking layer in the policy ensures our agent does not explore irrelevant areas.

In Figure \ref{fig:trajectory_example}, we show a comparison between the trajectories of $3$ agents (\textbf{Snp}, \textbf{BC} and \textbf{RL+BC+CL}). It is evident that the \textbf{BC} algorithm manages to mimic the \textbf{SnP} heuristic well, and that, although it never trained on this episode, it rarely chooses unwanted decisions (generalization capabilities). The \textbf{RL+BC+CL} trajectory has less actions and manages to grade more soil at an earlier step (see left column). Moreover, this policy reaches an overall higher reward and lower final height.

\paragraph{\textbf{Ablation Study}}\label{Ablation_study}
Our ablation study considered various network architectures, Gaussian masking coefficients and down-sampling. We expand on each architecture below. All these parameters are unit-less, as they were all normalized by some constant factor to allow a fair comparison.

\textbf{Network architecture}: We examined two architectures for the policy and value neural network: Resnet18 \cite{He_2016} and Yolo lite \cite{Huang_2018}. Ablations show that for the policy network, Yolo lite outperforms Resnet18, whereas for the value network, Resnet18's performance was superior (\ref{Tab:table_2}).

\begin{figure} % hbt!
     \centering
     \begin{subfigure}[b]{0.105\textwidth}
         \centering
         \includegraphics[width=\textwidth]{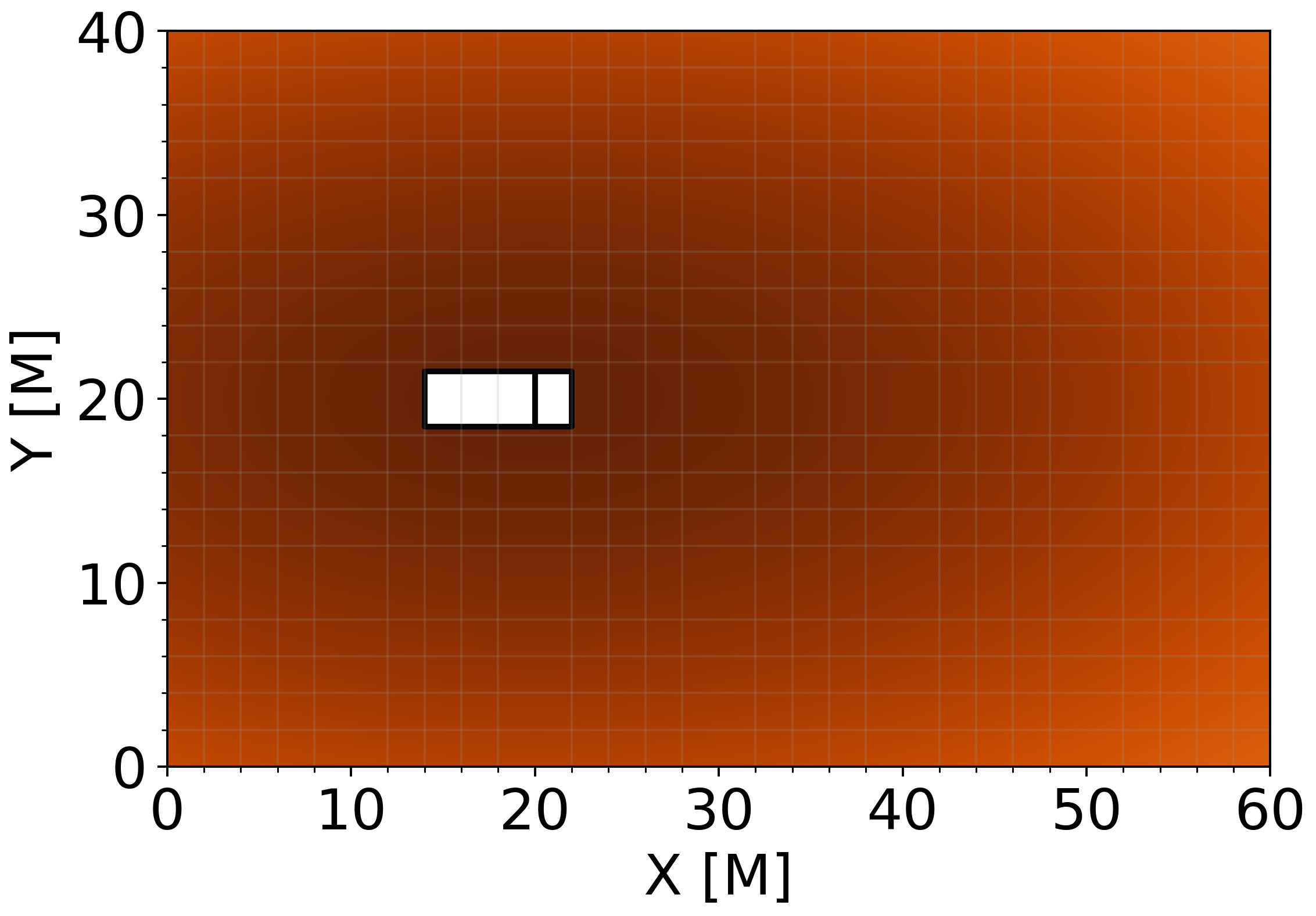}
         \caption{$\lambda_\sigma=1$}
         \label{fig:sigma_coeff1}
     \end{subfigure}
     \begin{subfigure}[b]{0.105\textwidth}
         \centering
         \includegraphics[width=\textwidth]{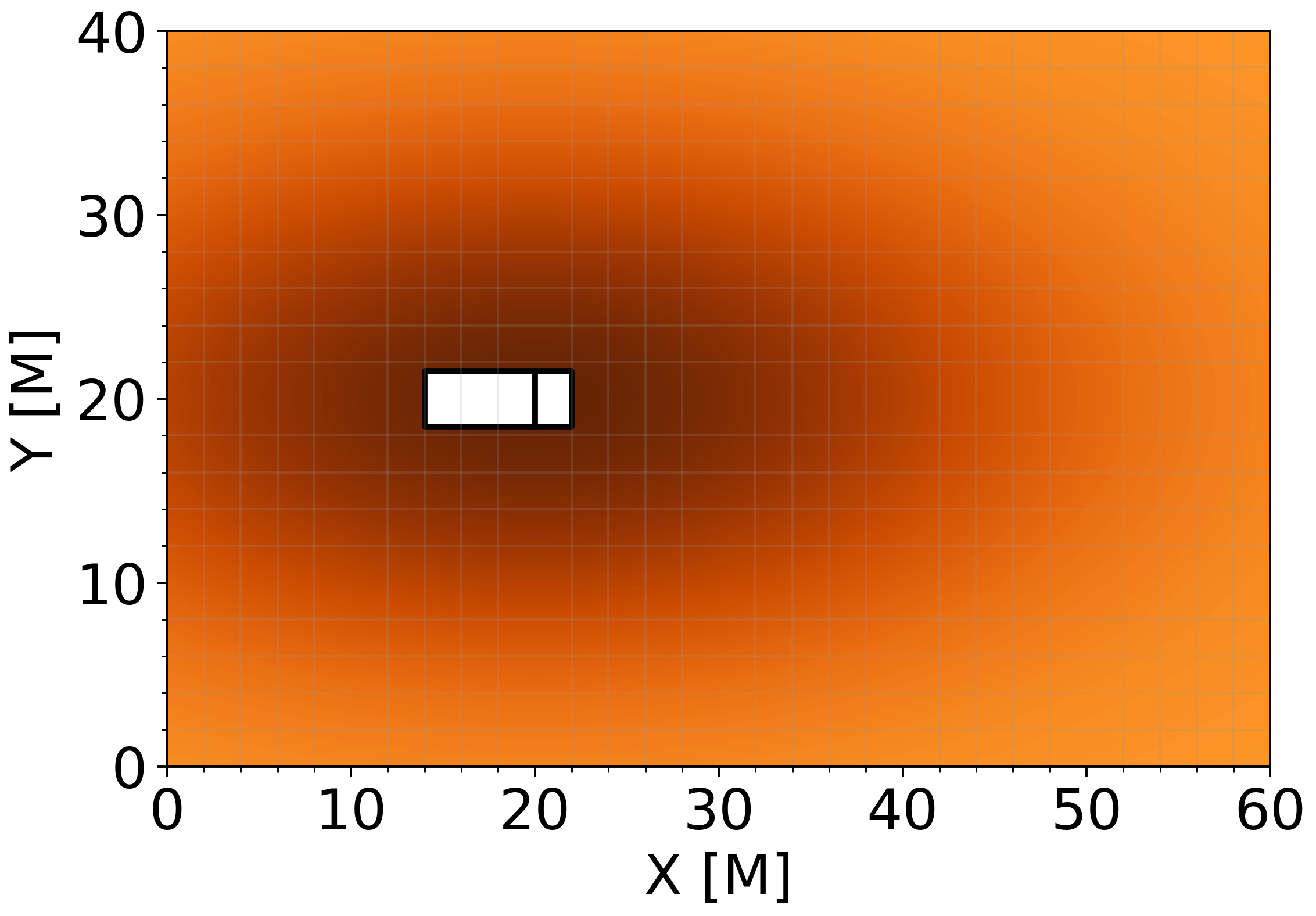}
         \caption{$\lambda_\sigma=2$}
         \label{fig:sigma_coeff2}
     \end{subfigure}
    \begin{subfigure}[b]{0.105\textwidth}
         \centering
         \includegraphics[width=\textwidth]{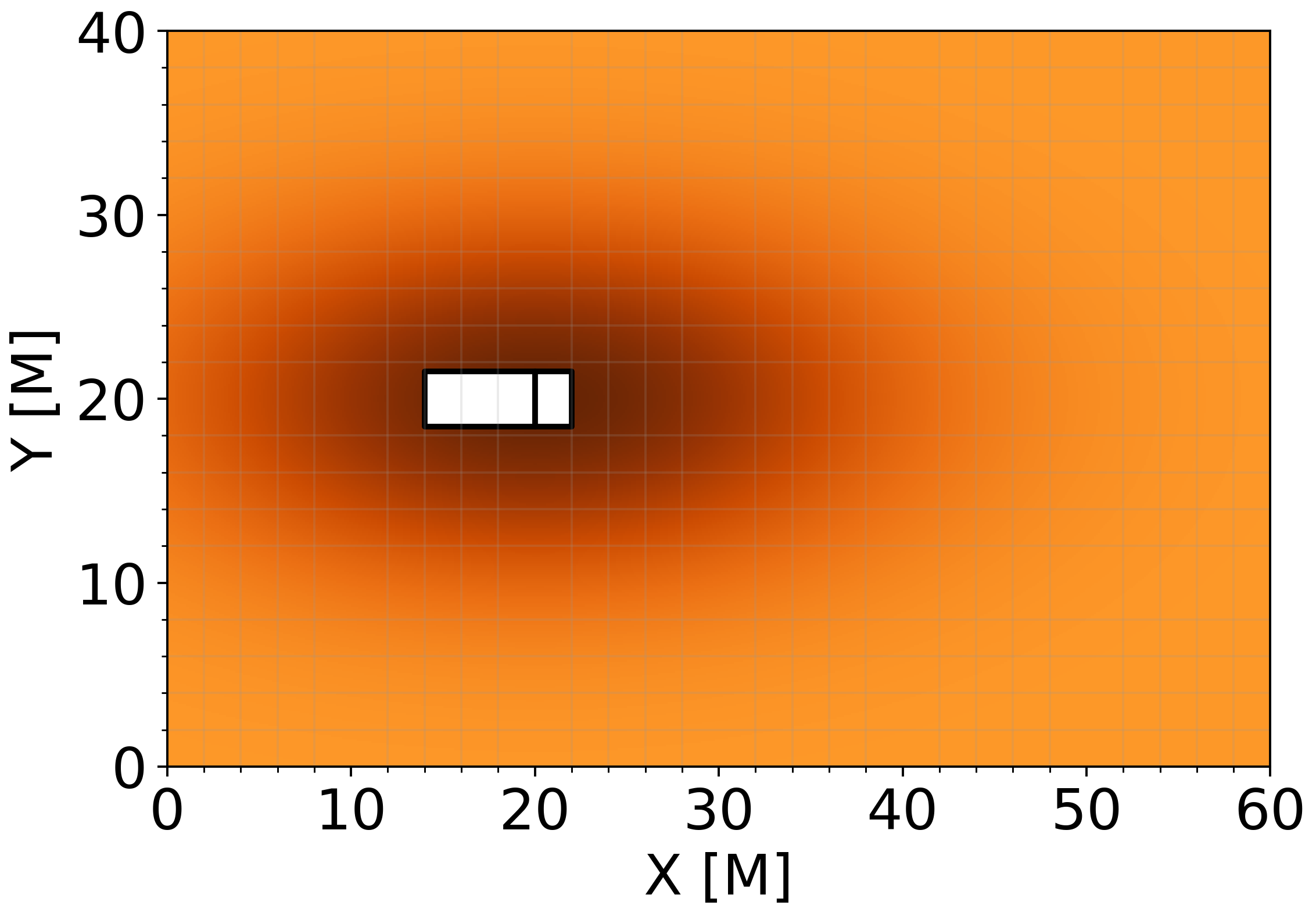}
         \caption{$\lambda_\sigma=3$}
         \label{fig:sigma_coeff3}
     \end{subfigure}
     \begin{subfigure}[b]{0.12\textwidth}
         \centering
         \includegraphics[width=\textwidth]{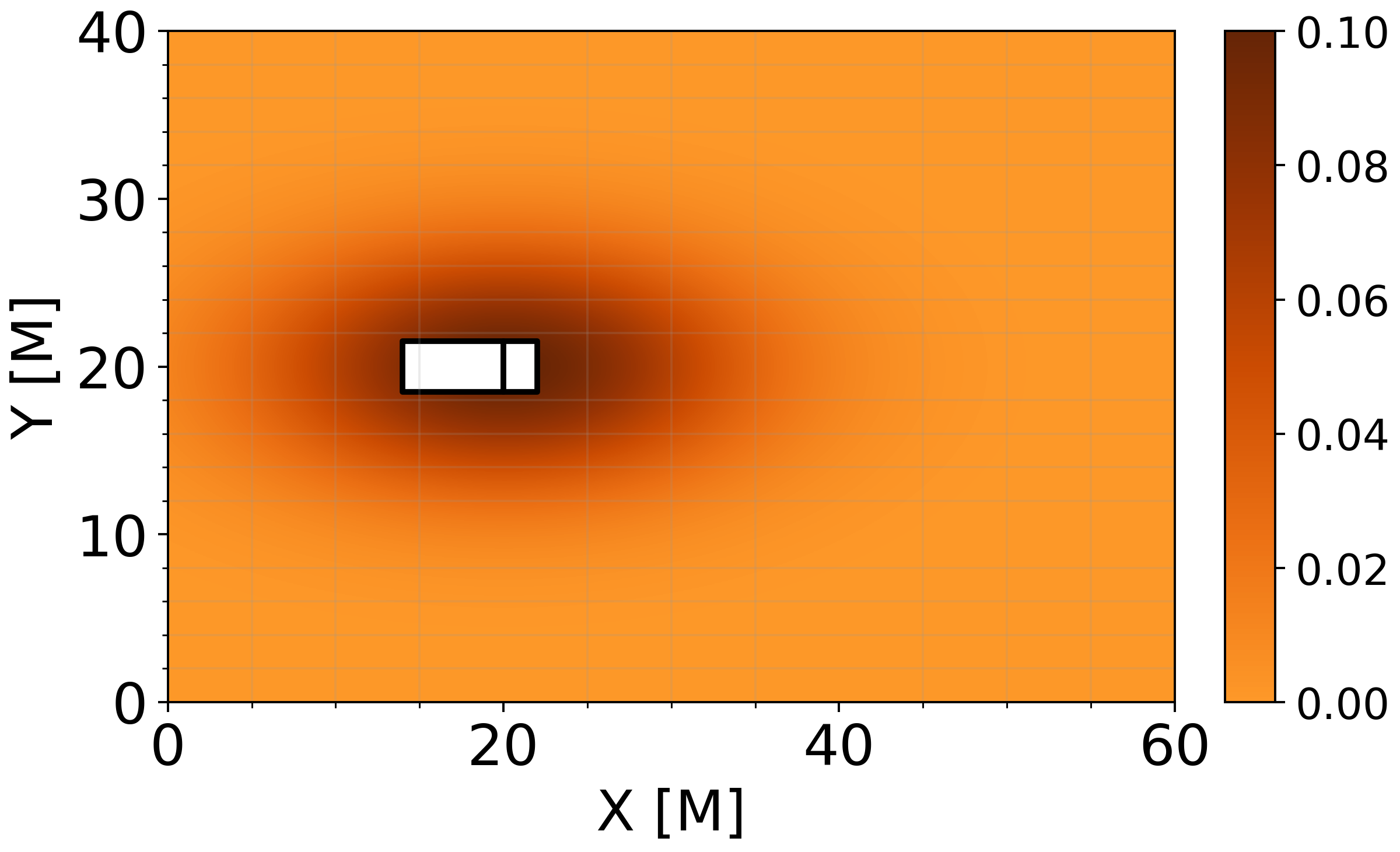}
         \caption{$\lambda_\sigma=4$}
         \label{fig:sigma_coeff4}
     \end{subfigure}
        \caption{Gaussian mask for different coefficients of $(\sigma_x, \sigma_y)$. Higher numbers indicate pixels with a high probability of being chosen}
        \label{fig:sigma_coeff_all}
\end{figure}

\textbf{Gaussian masking}: \label{Gaussian_masking} We added to the action space selection a Gaussian masking that allows the system to learn that actions that are closer to the current location are more beneficial, in terms of reward. 
% Without this mask, actions further away from the current location were selected, leading to an increase in the episode time and to sub-optimal performance. 
We examined \textit{four} values of the sigma factor $(\lambda_{\sigma})$ that was applied to generate the mask. A Gaussian sigma factor is the divisor of $(\sigma_x,\sigma_y)$ in the Gaussian mask. $(\lambda_{\sigma_x}, \lambda_{\sigma_y}) = (1,2,3,4)$, where "4" masks out pixels further away in the action space and "1" alleviates this restriction (see fig. \ref{fig:sigma_coeff_all} and  \ref{Tab:table_3}). Our results found the optimal value to be $3$.

\textbf{Down-sampling}: \label{down-sampling} 
As the selected action is sampled out of the learnt distribution of the pixels in the agent's FOV, we examined lowering the state space resolution to decrease both the memory footprint and the task time. Four values were examined, $2^{(1,2,3,4)}$, on the RL+BC+CL problem. Table \ref{Tab:table_1} depicts the results when the parameter was set to $2^3$. This configuration achieves the best results for removing the mean volume and height as well as the best rewards from the simulation environment. 
The number of parameters in the policy network is fairly small w.r.t. its performance. 
%  \begin{figure}
 \vspace{-1cm}
\begin{table}[hbt!] %hbt!
\begin{tabular}{|l|l|l|l|l|}
\hline
down-sample factor {[}pix{]} & $2^1$ & $2^2$ & $2^3$ & $2^4$ \\ \hline
state space size             & $300^2$     & $150^2$ & $75^2$  & $38^2$     \\ \hline
% \textbf{M} params  in policy network (yolo) $\downarrow$  & 3942     & 233     & 14.6     & 0.76     \\ \hline
params in policy net ($*1e6$) $\downarrow$  & $3942$     & $233$     & $14.6$     & $0.76$     \\ \hline

volume left [ ] $\downarrow$            & 8.1    &  2.8      & \textbf{1.3}     & 11.1     \\ \hline
total reward [ ] $\uparrow$              & 1385     & 3002     & \textbf{4441}     & 2992     \\ \hline
mean height left [ ] $\downarrow$       & 22     & 8     & \textbf{4}     & 27     \\ \hline
\end{tabular}
\caption{State-space down-sampling parameter}
\label{Tab:table_1}
\end{table}

% \begin{table}[!ht]
\begin{table}[hbt!]  %hbt!
\begin{tabular}{|c|r|r|r|r|}
\hline
\multicolumn{1}{|l|}{ablation net} & \multicolumn{1}{l|}{\begin{tabular}[c]{@{}l@{}}P=r, V=r\end{tabular}} & \multicolumn{1}{l|}{\begin{tabular}[c]{@{}l@{}}P=r, V=Y\end{tabular}} & \multicolumn{1}{l|}{\begin{tabular}[c]{@{}l@{}}P=Y, V=r\end{tabular}} & \multicolumn{1}{l|}{\begin{tabular}[c]{@{}l@{}}P=Y, V=Y\end{tabular}} \\ \hline
volume left {[} {]} $\downarrow$        & 71   & 21   & \textbf{4}     & 18 \\ \hline
total reward {[} {]} $\uparrow$         & 7245 & 5630 & \textbf{10149} & 6601 \\ \hline
mean height left {[} {]} $\downarrow$   & 239  & 61   & \textbf{19}    & 51 \\ \hline
\end{tabular}
\caption{networks architecture ablation, P means policy, V means value. r means Resnet, Y means Yolo}

\label{Tab:table_2}
\end{table}

% \begin{table}[!ht]
\begin{table}[hbt!]  %

\begin{tabular}{|c|r|r|r|r|}
\hline
\multicolumn{1}{|l|}{Gaussian scale factor ablation} & \multicolumn{1}{l|}{\begin{tabular}[c]{@{}l@{}}1\end{tabular}} & \multicolumn{1}{l|}{\begin{tabular}[c]{@{}l@{}}2\end{tabular}} & \multicolumn{1}{l|}{\begin{tabular}[c]{@{}l@{}}3\end{tabular}} & \multicolumn{1}{l|}{\begin{tabular}[c]{@{}l@{}}4\end{tabular}} \\ \hline
volume left {[} {]} $\downarrow$        & 8.4   & 2.6   & \textbf{2.2}     & 11.3 \\ \hline
total reward {[} {]} $\uparrow$         & 2375 & 2620 & \textbf{3987} & 2191 \\ \hline
mean height left {[} {]} $\downarrow$   & 22.2  & 7.4   & \textbf{6.5}    & 30.1 \\ \hline
\end{tabular}
\caption{Gaussian masking ablation}

\label{Tab:table_3}
\end{table}

\vspace{1cm}
\section{Discussion} \label{Discussion}
We here present AGPNet, an end-to-end pipeline for autonomous grading using a dozer.
First, we formulate the problem as an MDP and use RL and BC algorithms to solve the problem. 
Second, we create a light yet detailed simulation for training algorithms and suggest a new and innovative approach for simulating earth-moving vehicles and their interaction with the soil. 
We prove the validity of our simulation with a real prototype dozer and show how height-maps from the real dozer are comparable to the ones from our simulation.
Last, we train multiple policies and show that combining different RL and BC approaches with a high level of detection training such as CL achieves on par results with the heuristic and generalizes in more complex scenarios.
Our method is ideal for tasks where a vehicle has an interaction with the soil that effects the environment and changes the optimal sequence of  actions. It can also be used in other construction vehicles where the way-point planning is complex but the low-level actions can be %
defined using simple rule-based methods.
\section{Acknowledgements}
This work is part of a joint project between \href{https://www.bosch-ai.com}{Bosch-AI} and \href{https://www.shimz.co.jp/en/}{Shimizu} aimed at making autonomous grading agents available, reliable and robust.

% \bibliographystyle{IEEEtran}
\bibliographystyle{plain}
\bibliography{ICRA_Dozer}

\end{document}